\pdfoutput=1
\documentclass{article}

\usepackage[
  paperheight=8.5in,
  paperwidth=5.5in,
  left=10mm,
  right=10mm,
  top=20mm,
  bottom=20mm]{geometry}
\usepackage[utf8]{inputenc}

\usepackage{graphicx}
\usepackage{wrapfig}
\usepackage[bottom]{footmisc}
\usepackage{listings}
\usepackage{enumitem}

\usepackage{wrapfig}
\usepackage{ragged2e}

\usepackage{array}
\usepackage[table]{xcolor}
\usepackage{multirow}
\usepackage{booktabs}
\usepackage{hhline}
\definecolor{palegreen}{rgb}{0.6,0.98,0.6}

\usepackage{amsmath}
\usepackage{amssymb}
\usepackage{multicol}
\usepackage{lipsum}
\usepackage{hyphenat}
\PassOptionsToPackage{hyphens}{url}
\usepackage{url}

\usepackage{rotating}

\usepackage[T1]{fontenc}
\usepackage{textcomp}
\usepackage{listings}
\usepackage{setspace}

\newcommand*{\email}[1]{\small{\texttt{#1}}}

\renewcommand{\footnoterule}{%
  \kern -3pt
  \hrule width \textwidth height 0.5pt
  \kern 2pt
}

\date{}

\usepackage{titlesec}
\titleformat*{\section}{\large\bfseries}
\titleformat*{\subsection}{\normalsize\bfseries}
\titleformat*{\subsubsection}{\normalsize\bfseries}

\usepackage{biblatex}

\title{Bridging Traditional Machine Learning and Large Language Models: A Two-Part Course Design for Modern AI Education}

\author{
  Fang Li\\
  Computer Science Department\\
  Oklahoma Christian University\\
  Edmond, 73013\\
  \email{fang.li@oc.edu}
}

\begin{document}
\maketitle

\begin{abstract}
This paper presents an innovative pedagogical approach for teaching artificial intelligence and data science that systematically bridges traditional machine learning techniques with modern Large Language Models (LLMs). We describe a course structured in two sequential and complementary parts: foundational machine learning concepts and contemporary LLM applications. This design enables students to develop a comprehensive understanding of AI evolution while building practical skills with both established and cutting-edge technologies. We detail the course architecture, implementation strategies, assessment methods, and learning outcomes from our summer course delivery spanning two seven-week terms. Our findings demonstrate that this integrated approach enhances student comprehension of the AI landscape and better prepares them for industry demands in the rapidly evolving field of artificial intelligence.
\end{abstract}

\section{Introduction}

The artificial intelligence landscape has undergone unprecedented transformation with the emergence of Large Language Models (LLMs). While foundational machine learning techniques remain essential, the capabilities demonstrated by models such as GPT \cite{brown2020language}, LLaMA \cite{touvron2023llama}, and Claude \cite{anthropic2023claude} have fundamentally altered how AI systems are conceptualized, developed, and deployed. This paradigm shift presents a critical pedagogical challenge: how can educators effectively integrate established machine learning fundamentals with contemporary LLM technologies within a coherent curriculum framework?

Existing AI and data science courses typically adopt one of two approaches: those emphasizing classical statistical methods and traditional ML algorithms, or those focusing on neural networks and deep learning architectures. However, few successfully bridge these foundational approaches with modern LLM paradigms. This pedagogical gap can result in graduates who either possess strong theoretical foundations but lack practical experience with current tools, or who demonstrate familiarity with cutting-edge models but have gaps in fundamental understanding of underlying principles.

To address this educational challenge, we developed a comprehensive course structured as two sequential yet interconnected components: (1) pre-LLM foundations encompassing traditional machine learning and data science concepts, and (2) LLM applications and development covering contemporary AI technologies. This paper describes our pedagogical approach, implementation methodology, assessment strategies, and preliminary outcomes from delivering this curriculum as an intensive summer course spanning two seven-week terms.


\section{Related Work}

Traditional approaches to AI education have followed well-established frameworks. Classical AI courses typically adhere to the comprehensive structure outlined by Russell and Norvig \cite{russell2010artificial}, covering search algorithms, knowledge representation, reasoning systems, and introductory machine learning concepts. Specialized machine learning curricula generally emphasize statistical foundations and algorithmic approaches as described by Mitchell \cite{mitchell1997machine}, with contemporary extensions incorporating deep learning methodologies \cite{goodfellow2016deep}.

Recent educational developments have produced courses specifically targeting deep learning \cite{ng2018deep} and natural language processing \cite{jurafsky2000speech}. Becker et al. \cite{becker2023programming} describe the integration of transformer-based AI code generation tools into computing education, while Holland-Minkley et al. \cite{holland2023computer} examine challenges associated with teaching rapidly evolving AI technologies in liberal arts contexts. However, integrated ML–LLM approaches remain rare in current literature.

Our work extends these foundations by proposing a systematic two-part structure that explicitly connects traditional machine learning concepts with contemporary LLM paradigms, incorporating collaborative project-based learning and professional software development practices throughout the curriculum.

\section{Course Design Philosophy and Architecture}

\subsection{Two-Part Curriculum Structure}

The course employs a deliberate division into two distinct yet complementary components designed to create conceptual bridges between traditional and contemporary AI approaches:

\textbf{Part 1: Foundational Machine Learning} establishes core competencies in traditional data science and machine learning methodologies, including data preprocessing and exploration, feature engineering techniques, model selection and evaluation frameworks, and fundamental algorithmic approaches. This foundation ensures students develop solid understanding of the mathematical and statistical principles underlying all machine learning applications.

\textbf{Part 2: Large Language Model Applications} introduces students to contemporary LLM architectures, capabilities, and limitations while covering practical implementation aspects including prompt engineering strategies, retrieval-augmented generation systems, model deployment techniques, fine-tuning methodologies, and agent-based systems using Model Context Protocol.

This sequential structure enables students to appreciate how modern LLMs extend and build upon traditional machine learning concepts rather than viewing them as entirely separate technological paradigms. Students develop the ability to select appropriate tools and techniques based on problem requirements and constraints.

\subsection{Pedagogical Framework}

Our instructional approach is guided by four core pedagogical principles:

\textbf{Theory-Practice Integration} ensures each topic combines rigorous theoretical explanations with hands-on implementation exercises, enabling students to understand both conceptual foundations and practical applications. Rather than treating theory and practice as separate components, we integrate them throughout each learning module.

\textbf{Scaffolded Complexity} structures the progression from fundamental concepts to advanced applications, allowing students to build knowledge incrementally. Each new concept builds on previously established foundations.

\textbf{Authentic Applications} grounds examples and exercises in real-world use cases and industry scenarios, helping students understand the practical relevance and applicability of their learning. This approach connects academic concepts to professional practice and career preparation.

\textbf{Technological Accessibility} implements all course materials using Google Colab notebooks, eliminating barriers related to specialized hardware requirements and ensuring equitable access for all students regardless of their technological resources.

\section{Detailed Course Content and Implementation}

\subsection{Part 1: Foundational Machine Learning}

The first component establishes essential competencies across four integrated modules:

\textbf{Data Science Fundamentals and Exploratory Analysis} introduces students to the data science workflow through comprehensive analysis of the Titanic dataset. Students learn to identify patterns and relationships, handle missing data systematically, and create informative visualizations that reveal connections between variables such as passenger demographics and survival outcomes.

\textbf{Feature Engineering and Data Preprocessing} builds directly on initial analysis by teaching students to transform raw data into features suitable for machine learning applications. Students create derived variables (such as family size indicators), implement categorical encoding schemes, normalize numerical features, and evaluate the impact of preprocessing decisions on model performance. This module establishes the critical connection between domain knowledge and algorithmic success.

\textbf{Model Development and Evaluation} guides students through implementing multiple algorithmic approaches (including logistic regression, random forests, and support vector machines) on consistent datasets, enabling direct performance comparisons. Students practice cross-validation techniques, interpret confusion matrices and ROC curves, and develop skills in selecting appropriate evaluation metrics for different problem types. This module emphasizes the importance of rigorous experimental methodology.

\textbf{Advanced Machine Learning Concepts} introduces neural network architectures through both theoretical foundations and practical implementation. Students build feedforward networks for classification tasks, explore convolutional neural networks using the MNIST dataset, and implement recurrent neural networks for sequence prediction. This module creates explicit connections to the transformer architectures that underpin modern LLMs.

\subsection{Part 2: Large Language Model Applications}

The second component transitions students to contemporary AI technologies through six integrated modules:

\textbf{Large Language Model Foundations} explores the evolution from traditional NLP approaches to modern transformer-based architectures. Students examine attention mechanisms through interactive visualizations, implement simplified transformer components, and compare outputs across different model scales to understand emergent capabilities. This module explicitly connects to neural network concepts from Part 1.

\textbf{LLM Platforms and Development Tools} provides hands-on experience with the Hugging Face ecosystem and LangChain framework. Students build question-answering systems using pre-trained models, create processing chains for complex applications, and develop familiarity with industry-standard tools and libraries. This module emphasizes practical development skills.

\textbf{Prompt Engineering and Retrieval-Augmented Generation} teaches students to optimize model performance through strategic prompt design and external knowledge integration. Students experiment with various prompting techniques, analyze their effectiveness across different tasks, and implement retrieval-augmented generation systems using vector databases. This module demonstrates how traditional information retrieval concepts enhance LLM capabilities.

\textbf{Model Deployment and Optimization} covers practical aspects of making models accessible to end users, including quantization techniques for resource optimization, web interface development using Gradio, and API creation with FastAPI. 

\textbf{Model Customization and Fine-tuning} enables students to adapt pre-trained models for specific domains and applications. Students compare full fine-tuning approaches with parameter-efficient techniques such as LoRA, analyzing trade-offs in performance, resource requirements, and training efficiency. This module connects to model training concepts from Part 1 while addressing contemporary scaling challenges.

\subsection{Technical Implementation Strategy}

All course materials are implemented as Google Colab notebooks, providing several educational and practical advantages. The cloud-based environment ensures accessibility for students without powerful local hardware, creates consistency across student experiences, integrates code with explanations and visualizations in unified documents, and enables reproducible execution of all course examples. To manage Colab’s runtime limits, students use checkpoints and lightweight frameworks.

For LLM-related applications, we utilize efficient models such as DistilGPT-2, Phi-2, and quantized LLaMA-3.1, which operate within Colab’s resource constraints while teaching students about the capabilities and applications of larger production models. Students implement retrieval-augmented generation (RAG) systems using libraries like Hugging Face’s \texttt{transformers} and \texttt{sentence-transformers}, enabling them to create agents that retrieve relevant documents from simulated knowledge bases (e.g., a vector database of lecture notes using FAISS) and generate context-aware responses. This approach balances hands-on experience with practical resource management skills while introducing industry-standard techniques for data-driven agent development.

\section{Assessment Strategy and Learning Activities}

\subsection{Comprehensive Assessment Framework}

The course employs a multi-faceted assessment approach designed to evaluate both individual competency development and collaborative skills:

\textbf{Individual Assignments} (40\% of final grade) consist of nine targeted assignments, one per major topic, designed to reinforce lecture concepts and provide immediate practice with techniques. For example, following the feature engineering module, students implement various preprocessing approaches on novel datasets not previously used in class demonstrations. These assignments ensure individual accountability and concept mastery.

\textbf{Collaborative Group Projects} (40\% of final grade) engage students in two major team-based endeavors that integrate multiple course concepts. Students work in consistent groups of four throughout the entire course, fostering sustained collaboration and team development. The first project requires groups to develop conventional machine learning solutions for real-world problems, while the second project involves creating sophisticated LLM-based applications that integrate multiple contemporary techniques including prompt engineering, retrieval-augmented generation, and Model Context Protocol implementations for tool integration. This structure allows students to apply foundational concepts before building on them with advanced technologies.

\textbf{Active Participation} (10\% of final grade) includes class exercises, discussions, and notebook activities.

\textbf{Peer Collaboration Evaluation} (10\% of final grade) involves students assessing their group members' contributions, communication effectiveness, and collaborative skills. This component emphasizes the importance of professional teamwork and provides accountability for group dynamics.

\subsection{Professional Development Integration}

An essential component of the group projects involves learning industry-standard software development and project management practices:

\textbf{Version Control and Collaboration} requires students to use GitHub for all project work, including branch creation, pull request workflows, and systematic code review processes. Students learn to manage collaborative development and resolve conflicts in shared codebases.

\textbf{Documentation and Communication} mandates comprehensive project documentation including detailed README files explaining project purposes and usage, inline code comments clarifying complex logic, documentation of architectural decisions and design rationales, and thorough reporting of performance metrics and evaluation results.

\textbf{Project Management} involves creating and maintaining project boards using GitHub Projects or similar tools to track tasks, assign responsibilities, and manage deadlines. Groups must demonstrate systematic planning and progress monitoring throughout project development.

\textbf{Quality Assurance} includes peer code review within groups, ensuring that all team members review and provide feedback on contributions from colleagues. This process emphasizes code quality, knowledge sharing, and collaborative problem-solving.

\subsection{Interactive Learning Activities}

Each instructional module incorporates multiple hands-on exercises that enable immediate application of concepts. Representative activities include building text summarization tools using LangChain frameworks, optimizing model performance through quantization techniques, creating interactive chatbot interfaces with Gradio, fine-tuning language models on custom datasets, and implementing RAG-enabled agents that can retrieve relevant documents from simulated knowledge bases and generate context-aware responses. These exercises promote active learning and provide immediate feedback on comprehension and skill development.

\section{Learning Outcomes and Evaluation}

\subsection{Student Performance and Engagement}

Initial implementations of the course across multiple cohorts have demonstrated encouraging outcomes. Students consistently develop comprehensive understanding of the AI landscape, recognizing both the enduring value of traditional ML techniques and the transformative capabilities of modern LLMs. Final projects demonstrate creative integration of both paradigms, with students effectively combining traditional techniques for data preparation and analysis with LLMs for natural language understanding and generation tasks.

Student feedback indicates enhanced confidence for industry positions that increasingly require familiarity with both established ML methodologies and contemporary LLM technologies. Exit surveys reveal that students feel better prepared to evaluate when traditional approaches may be more appropriate than LLM solutions, and vice versa.

\subsection{Implementation Challenges and Solutions}

Several significant challenges emerged during course implementation, each requiring targeted pedagogical solutions:

\textbf{Heterogeneous Student Backgrounds} presented difficulties when some students entered with stronger programming or mathematical foundations than others. We addressed this challenge by developing supplementary resources for foundational skills, implementing peer mentoring systems within groups, and providing multiple pathways for concept explanation and practice.

\textbf{Computational Resource Limitations} arose from Colab's free tier constraints affecting work with larger models and datasets. We developed optimization strategies including model quantization techniques, efficient batching approaches, and alternative implementation methods that students could employ when facing resource constraints.

\textbf{Rapid Technological Evolution} required frequent updates to course materials as LLM technology continued advancing throughout course delivery. We restructured content around enduring fundamental concepts that remain relevant despite changes in specific models or tools, while maintaining flexibility to incorporate significant new developments.

\textbf{Conceptual Integration Difficulties} emerged when students struggled to connect traditional ML concepts with their LLM counterparts. We addressed this by developing explicit mapping exercises between concepts (such as feature engineering and prompt engineering) and creating assignment sequences that required students to solve similar problems using both approaches.

\section{Best Practices and Pedagogical Insights}

Through multiple course iterations, several key insights emerged that may benefit other educators developing similar curricula:

\textbf{Explicit Conceptual Bridging} proves essential for helping students understand connections between traditional ML and LLM paradigms. Students benefit significantly from structured discussions of how concepts like feature engineering relate to prompt engineering, or how traditional model evaluation connects to LLM performance assessment. We found success in creating comparison matrices and mapping exercises that make these connections explicit.

\textbf{Strategic Group Formation} yields better outcomes than random team assignment. We developed a skills assessment survey covering programming experience, mathematical background, and communication preferences, then formed balanced teams ensuring each group possessed complementary strengths. This approach reduced skill gaps within teams and improved collaborative dynamics.

\textbf{Resource Management Education} becomes increasingly important as students work with larger, more complex models. Teaching students to work effectively within computational constraints through techniques like model quantization, efficient batching, and strategic resource allocation proved essential for successful project completion in cloud-based environments.

\textbf{Integrated Ethics Education} works more effectively than standalone ethics modules. We embedded discussions of bias, fairness, transparency, and responsible AI throughout both course components, helping students develop consistent ethical frameworks that apply across all AI technologies rather than treating ethics as a separate concern.

\textbf{Peer Teaching Opportunities} significantly enhance learning outcomes. Having students present their solutions to in-class exercises, explain their project approaches to other groups, and teach concepts they've mastered to struggling peers reinforced learning and exposed the class to diverse perspectives and problem-solving approaches.

\section{Future Directions and Course Evolution}

As LLM technology and AI education continue evolving, we have identified several areas for course enhancement:

\textbf{Multimodal Model Integration} will expand coverage to include models that combine text, image, and audio capabilities, reflecting the increasing importance of multimodal AI systems in industry applications. This expansion will require developing new assessment methods and project frameworks that leverage multiple data modalities.

\textbf{Agent-Based Systems Development} will expand coverage of LLM-powered agents with Model Context Protocol (MCP), and include more complex multi-agent systems, workflow orchestration, and integration with enterprise-scale tool ecosystems. This addition reflects the growing importance of AI systems that can take coordinated actions across multiple domains rather than simply processing information in isolation.

\textbf{Responsible AI Integration} will strengthen focus on ethical considerations, bias mitigation strategies, and responsible deployment practices throughout both course components. This enhancement responds to growing industry and societal demands for AI practitioners who can navigate ethical challenges effectively.

\textbf{Industry Partnership Development} will involve creating connections with industry practitioners who can provide guest lectures, project mentorship, and real-world problem scenarios that reflect current professional challenges and opportunities.

\section{Conclusion}

The emergence of Large Language Models represents both a significant challenge and an unprecedented opportunity for computer science education. Our two-part course structure successfully addresses this challenge by creating systematic bridges between traditional machine learning foundations and contemporary LLM applications, ensuring students develop both conceptual depth and practical breadth in their AI education.

The key contributions of our pedagogical approach include the development of a structured progression that builds explicit conceptual connections between traditional ML and LLM paradigms, the integration of individual skill development with collaborative project-based learning that reflects industry practices, the incorporation of professional software engineering and project management practices through GitHub-based workflows, and the demonstration of balanced coverage between theoretical foundations and practical implementation techniques.

Our assessment of learning outcomes indicates that this integrated approach provides a flexible and robust framework that can adapt to continued technological developments while ensuring students master enduring principles that underpin all machine learning applications. By teaching students not only how to use contemporary AI technologies but also how to understand their foundations, evaluate their limitations, and select appropriate applications, we prepare them to become thoughtful and effective practitioners in an increasingly AI-driven professional landscape.

The course design demonstrates that it is possible to bridge traditional and contemporary approaches without sacrificing depth in either area. Students emerge with both the foundational knowledge necessary to understand new developments and the practical skills required to implement sophisticated AI solutions. This combination proves essential as the field continues evolving at an unprecedented pace.

Future work will focus on expanding curriculum coverage to include multimodal models and enhanced agent-based architectures, developing more sophisticated assessment methods for evaluating students' technology selection and application skills, and strengthening connections with industry partners to ensure continued relevance of course content and learning outcomes.

As AI technologies continue reshaping industries and society, educational approaches that successfully integrate foundational knowledge with contemporary applications become increasingly valuable. Our experience suggests that systematic, bridging approaches like the one described here can effectively prepare students for careers in this dynamic and rapidly evolving field.

\medskip

\printbibliography

@book{russell2010artificial,
  title={Artificial intelligence: a modern approach},
  author={Russell, Stuart J and Norvig, Peter},
  year={2010},
  publisher={Pearson}
}

@book{mitchell1997machine,
  title={Machine learning},
  author={Mitchell, Tom M},
  year={1997},
  publisher={McGraw-Hill}
}

@book{goodfellow2016deep,
  title={Deep learning},
  author={Goodfellow, Ian and Bengio, Yoshua and Courville, Aaron},
  year={2016},
  publisher={MIT press}
}

@misc{ng2018deep,
  title={Deep learning specialization},
  author={Ng, Andrew},
  year={2018},
  publisher={Coursera}
}

@book{jurafsky2000speech,
  title={Speech and language processing: An introduction to natural language processing, computational linguistics, and speech recognition},
  author={Jurafsky, Daniel and Martin, James H},
  year={2000},
  publisher={Prentice Hall}
}

@article{brown2020language,
  title={Language models are few-shot learners},
  author={Brown, Tom and Mann, Benjamin and Ryder, Nick and Subbiah, Melanie and Kaplan, Jared D and Dhariwal, Prafulla and Neelakantan, Arvind and Shyam, Pranav and Sastry, Girish and Askell, Amanda and others},
  journal={Advances in neural information processing systems},
  volume={33},
  pages={1877--1901},
  year={2020}
}

@article{touvron2023llama,
  title={LLaMA: Open and efficient foundation language models},
  author={Touvron, Hugo and Lavril, Thibaut and Izacard, Gautier and Martinet, Xavier and Lachaux, Marie-Anne and Lacroix, Timoth{\'e}e and Rozi{\`e}re, Baptiste and Goyal, Naman and Hambro, Eric and Azhar, Faisal and others},
  journal={arXiv preprint arXiv:2302.13971},
  year={2023}
}

@misc{anthropic2023claude,
  title={Claude: Anthropic's assistant for harmless, helpful conversations},
  author={Anthropic},
  year={2023},
  howpublished={\url{https://www.anthropic.com/claude}},
  note={Accessed: 2025-07-31}
}

@inproceedings{becker2023programming,
  title={Programming is hard-or at least it used to be: Educational opportunities and challenges of ai code generation},
  author={Becker, Brett A and Denny, Paul and Finnie-Ansley, James and Luxton-Reilly, Andrew and Prather, James and Santos, Eddie Antonio},
  booktitle={Proceedings of the 54th ACM Technical Symposium on Computer Science Education V. 1},
  pages={500--506},
  year={2023}
}

@inproceedings{holland2023computer,
  title={Computer science curriculum guidelines: A new liberal arts perspective},
  author={Holland-Minkley, Amanda and Barnard, Jakob and Barr, Valerie and Braught, Grant and Davis, Janet and Reed, David and Schmitt, Karl and Tartaro, Andrea and Teresco, James D},
  booktitle={Proceedings of the 54th ACM Technical Symposium on Computer Science Education V. 1},
  pages={617--623},
  year={2023}
}

\end{document}